\documentclass[letterpaper, 10 pt, conference]{ieeeconf}  

\IEEEoverridecommandlockouts                              

\usepackage{kotex}
\usepackage{amsmath}
\usepackage{amssymb}
\usepackage{array}
\usepackage{caption}
\usepackage{graphicx}
\usepackage{multirow}
\usepackage{tabularx}
\usepackage{threeparttable}
\usepackage{color}
\usepackage{adjustbox}
\usepackage{gensymb}
\definecolor{cobalt}{rgb}{0.0, 0.28, 0.9}

\title{\LARGE \bf
Category-level Neural Field for Reconstruction of \\Partially Observed Objects in Indoor Environment
}

\author{Taekbeom Lee, Youngseok Jang, and H. Jin Kim
\thanks{Taekbeom Lee and H. Jin Kim are with the Aerospace Engineering Department, Seoul National University, South Korea
        {\tt\small ltb1128@snu.ac.kr, hjinkim@snu.ac.kr}}%
\thanks{Youngseok Jang is with the Mechanical and Aerospace Engineering Department, Seoul National University, South Korea
        {\tt\small duscjs59@gmail.com}}%
\thanks{{This research was supported by Unmanned Vehicles Core Technology Research and Development Program through the National Research  Foundation of Korea(NRF) and Unmanned Vehicle Advanced Research Center(UVARC) funded by the Ministry of Science and ICT, the Republic of Korea(NRF-2020M3C1C1A010864)}}%
}

\begin{document}

\maketitle
\thispagestyle{empty}
\pagestyle{empty}


\begin{abstract}
Neural implicit representation has attracted attention in 3D reconstruction through various success cases. For further applications such as scene understanding or editing, several works have shown progress towards object-compositional reconstruction. Despite their superior performance in observed regions, their performance is still limited in reconstructing objects that are partially observed.
To better treat this problem, we introduce category-level neural fields that learn meaningful common 3D information among objects belonging to the same category present in the scene. Our key idea is to subcategorize objects based on their observed shape for better training of the category-level model. Then we take advantage of the neural field to conduct the challenging task of registering partially observed objects by selecting and aligning against representative objects selected by ray-based uncertainty.
Experiments on both simulation and real-world datasets demonstrate that our method improves the reconstruction of unobserved parts for several categories.
\end{abstract}

\section{INTRODUCTION}
\label{INTRODUCTION}

Recent approaches leveraging neural implicit representations have shown promising outcomes not only in view synthesis but also in 3D reconstruction. Particularly, NeRF employs a multi-layer perceptron (MLP) to train a continuous function mapping 3D coordinates to the associated volume density and radiance. Such coordinate-MLP methods have the advantage of compact memory representations over explicit representation, as they encode scene information through neural network parameters. Their continuous nature offers complete reconstructions, addressing challenges such as unobserved region holes often seen in classical volumetric fusion methods. 

\begin{figure}[t]
\centering
\includegraphics[width = 1\linewidth]{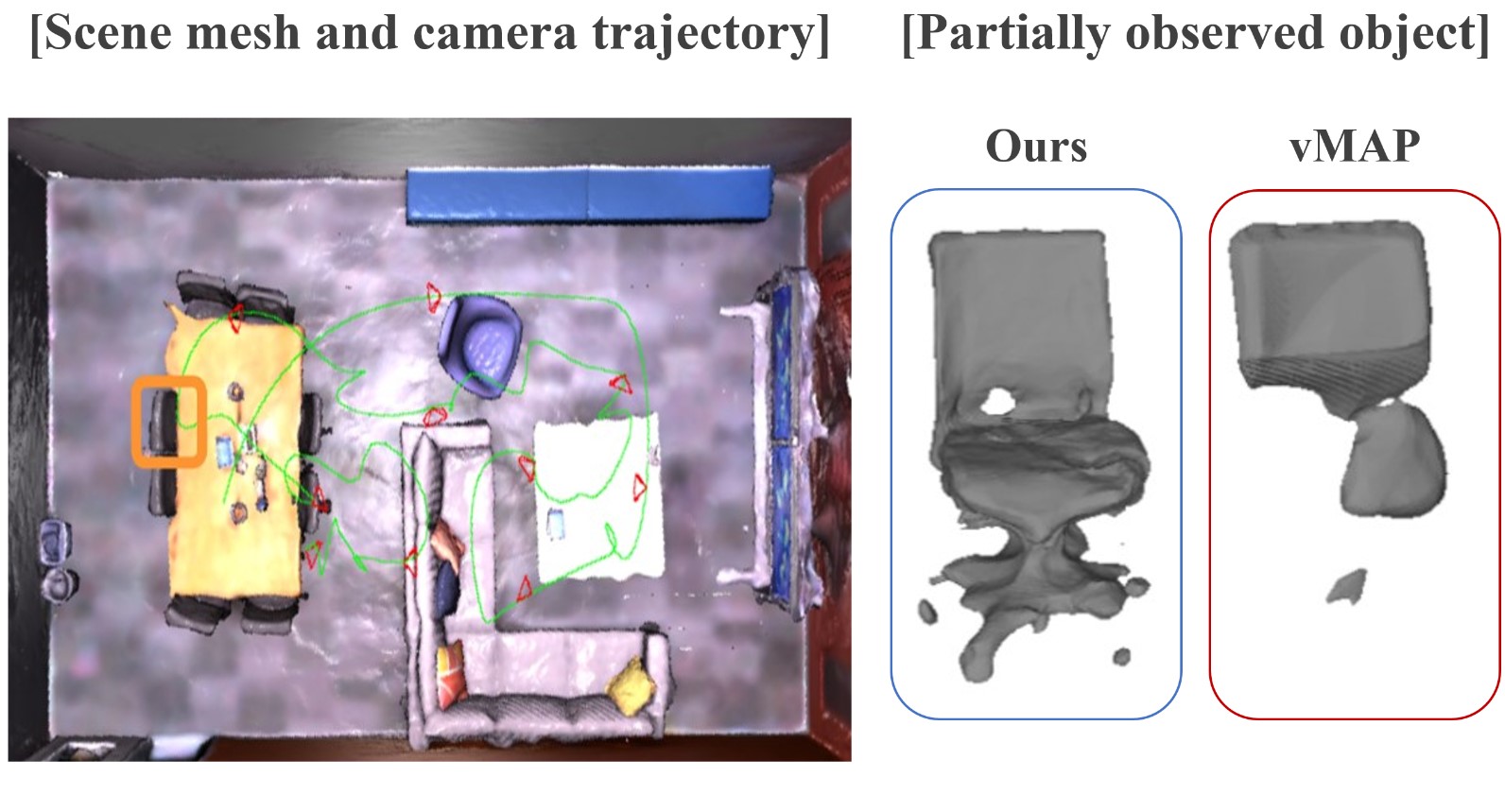}
\caption{Our method reconstructs objects using category-level models. Objects belonging to the same category share common shape properties, which help to reconstruct unobserved parts plausibly. On the other hand, unobserved parts of objects that are reconstructed by the object-level model (vMAP) tend to over-smooth or fail to recover complete geometry.}
\vspace{-5mm}
\label{fig:figure1}
\end{figure}

However, the majority of coordinate-MLP methods focus on scene-level representations, leaving a notable gap in achieving object-level understanding. Object-compositional representation decomposes scene into semantic units, objects, and this semantic composition further advances potential applications such as scene understanding \cite{nie2020total3dunderstanding}\cite{li2021moltr}, editing \cite{yang2021learning}\cite{ost2021neural}, and AR/VR \cite{yu2021unsupervised}\cite{wu2023object}. Some research has proposed NeRF-based models capable of representing both the scene and individual objects using 2D object masks as additional input. Some of them train a single MLP which represents the whole scene and leverages a specific branch to represent objects \cite{yang2021learning}\cite{wu2022object}. However, these methods are inefficient in that the entire network must be learned and inferred to represent a single object.

Other studies train object-wise MLP. Most of these methods are category-level, which leverage separate MLP for each category to learn common characteristics such as the shape and texture of objects in the same category \cite{ost2021neural}\cite{jang2021codenerf}\cite{muller2022autorf}. The learned model acts as prior knowledge for its category to reconstruct objects which was only partially observed during training. However, these methods can be only used for limited categories that have sufficiently large amounts of data. Some studies \cite{kong2023vmap}\cite{kundu2022panoptic} train instance-wise MLP and overcome this limitation. vMAP \cite{kong2023vmap} models the neural field of objects into separate neural networks trained independently and shows it can reconstruct watertight object mesh without prior. Panoptic Neural Field \cite{kundu2022panoptic} also incorporates separate neural fields for individual objects and employs meta-learned initialization as a category-level prior only if a large dataset for the category exists. 
However, they do not utilize common information intra-category, and their performance in reconstructing unseen parts of objects remains insufficient without prior.

We accordingly propose a category-level model-based method that does not use prior knowledge while utilizing common information in the category. We train NeRF model for each category of observed objects. We do not directly use the output of semantic segmentation as each object's category because semantic segmentation models predict an object's category mainly considering the semantic commonality, and even objects in the same category can significantly differ in shape. Instead, we estimate shape similarity between each object pair and break objects with different shapes into subcategories. To train common knowledge shared by the objects in the same subcategory, it is necessary to align the objects in 3D space. Since the observed parts vary for each partially observed object, aligning objects presents a challenge in learning category-level models. To address this challenge, we determine the most well-observed object as representative for each category using a ray-based uncertainty metric and transform other objects into their normalized object coordinate space (NOCS). Experiments on both simulated and real-world datasets show that our method can improve the reconstruction of unobserved parts of common objects in indoor scenes. In summary, the primary contributions of the paper are:

\begin{itemize}
    \item We propose an object-level mapping system that enhances reconstruction capabilities for unobserved parts by learning category-level knowledge solely from the observed data.
    \item We propose a method that adaptively subcategorizes objects based on their observed shape, which allows the objects to share common 3D information through category-level models.
    \item We propose to decide representative using ray-based uncertainty and register objects to its NOCS per each category, which addresses the challenge of learning category-level models from partially observed objects. 
\end{itemize}
\begin{figure*}[t]
\centering
\includegraphics[width = .9\linewidth]{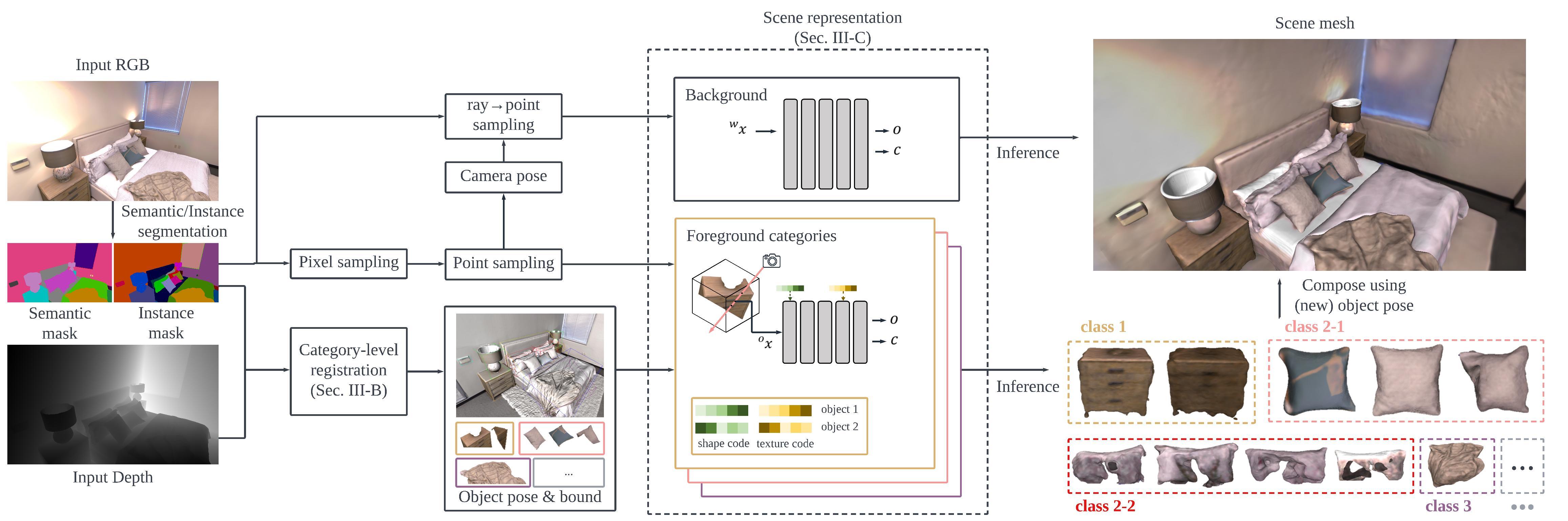}
\caption{Overview of the proposed framework}
\vspace{-7mm}
\label{fig:pipeline}
\end{figure*}
\section{RELATED WORK}
\label{related work}
\subsection{Neural implicit representation for 3D reconstruction} 
Recent years have seen significant interest in studies that utilize neural network to learn implicit representations for 3D shape reconstruction. They adopt a variety of representations, such as occupancy \cite{mescheder2019occupancy}\cite{peng2020convolutional}\cite{niemeyer2020differentiable}, signed distance functions \cite{park2019deepsdf}\cite{yariv2021volume}\cite{wang2021neus}, and density \cite{mildenhall2020nerf}. Of particular interest is NeRF \cite{mildenhall2020nerf} which shows impressive results in novel view synthesis. NeRF employs a multi-layer perceptron (MLP) to represent a scene using a neural radiance field comprised of volume density and radiance, and leverage classical volumetric rendering. Its capability to effectively capture geometric details, in addition to its capability for complete reconstruction and memory efficiency, has led to its application not only for single object but also in room-level dense 3D reconstructions \cite{yu2022monosdf}\cite{azinovic2022neural}\cite{wang2022go}. Some works \cite{sucar2021imap}\cite{zhu2022nice}\cite{rosinol2022nerf}\cite{wang2023co} achieve real-time SLAM utilizing active sampling, keyframe selection, or applying frameworks\cite{muller2022instant}\cite{chen2022tensorf} for accelerating training time. All of these methods consider representing the entire scene only, and they are not capable of reconstructing individual object and further applications such as object manipulation.

\subsection{Object-compositional neural implicit representation}
\textbf{Scene-level methods.}
Following promising results obtained in 3D reconstruction with neural implicit representations, efforts have been made to acquire individual object representations using neural fields. Most studies model all objects present in a scene using a single MLP \cite{yang2021learning}\cite{wu2022object}\cite{li2023rico}\cite{wu2023objectsdf++}. 
ObjectNeRF \cite{yang2021learning} uses an additional object branch which renders individual objects. ObjSDF \cite{wu2022object} predicts each object's SDF from a unified branch and renders both scene and individual objects based on this prediction. However, these methods are inefficient in both learning and rendering, as a large amount of network parameters are shared, even among dissimilar objects.

\textbf{Category-level methods.}
Category-level NeRF methods train and render various objects within a category using a single NeRF model combined with individual latent codes \cite{jang2021codenerf}\cite{muller2022autorf}\cite{henzler2021unsupervised}\cite{xie2021fig}. 
Each model learns category-level prior knowledge from the training dataset and successfully renders even in challenging situations such as few-shot scenarios at test time. CodeNeRF \cite{jang2021codenerf} conditions separate shape and texture codes for each instance on a shared MLP within a category, achieving disentanglement of shape and texture for individual objects. AutoRF \cite{muller2022autorf} adopts an encoder-decoder model structure that encodes shape and texture from images and uses category-level NeRF as the decoder. However, these approaches require extensive training with a large number of objects within each category, limiting their application to specific categories. In contrast, our method learns category-level information existing in observed scenes for \textit{any} category, eliminating the need for external datasets.

\textbf{Instance-specific methods.}
Alternatively, some studies have employed independent neural fields for each object \cite{kong2023vmap}\cite{kundu2022panoptic}. Panoptic Neural Field \cite{kundu2022panoptic} trains with a separate NeRF for the background and each object. vMAP \cite{kong2023vmap} also models each object with a neural field, independently learning for each object using its own keyframe buffer to perform efficient online object mapping. However, vMAP struggles to reconstruct unseen parts of objects successfully because it does not learn common characteristics within a category. In contrast, our approach learns shared properties of a category from different objects within the scene, enabling more effective reconstruction of occluded regions.

\section{METHOD}
We reconstruct individual objects and composite a complete indoor scene using $M$ posed RGB-D sequences $\mathcal{I} = \left\{ I_{i}\right\}_{i=\{1,\ldots,M\}}$ and $\mathcal{D} = \left\{ D_{i}\right\}_{i=\{1,\ldots,M\}}$. For each frame, high-quality instances and semantic segmentation masks are given from an off-the-shelf 2d instance segmentation network or the dataset itself. The overview of our method is shown in Fig. \ref{fig:pipeline}.

\subsection{Preliminaires}
\label{preliminaries}
\textbf{NeRF.}
NeRF learns neural implicit representation from multi-view images. Specifically, it takes posed images as input and represents a scene in terms of volume density and radiance. NeRF employs an MLP to learn an implicit function $f$ which maps a 3d point $p\in \mathbb{R}^{3}$ and a viewing direction $d\in \mathbb{R}^{3}$ to volume density $\sigma \in \mathbb{R}$ and color $c\in \mathbb{R}^{3}$ for the given scene. To render a 2D image, NeRF casts a ray $r(t)=o+td \; (t\geq0)$ from the camera origin $o$ in the direction $d$ towards each pixel. The radiance of each pixel is approximated by integrating the radiance along the ray.
\begin{align}
C(r) = \int_{0}^{\infty}T(t)\sigma(r(t))c(r(t),d)dt\;,
\end{align}
where $T(t)$ represents the accumulated transmittance along the ray, meaning the probability that the ray travels to $t$ without any collision with other particles. This formulation can be approximated as a weighted sum through the quadrature rule \cite{max1995optical}. Similarly, depth can be also rendered:
\begin{align}
    \hat{C}(r) = \sum_{i=1}^{N}w_{i}c_{i}, \quad \hat{D}(r) = \sum_{i=1}^{N}w_{i}d_{i}\;,
\end{align}
where $w_{i}=T_{i}(1-\exp(-\sigma_{i}\delta_{i}))$, $T_{i} = \exp(-\sum_{j=1}^{i-1}\sigma_{j}\delta_{j})$, $\delta_{i}=t_{i+1}-t_{i}$, and $N$ is the number of samples along the ray. 

\textbf{Object surface rendering.}
For each object given its 2d instance mask, a separate neural field is trained using rays that travel through pixels inside the 2d bounding box of the object \cite{kong2023vmap}. To get a clear boundary of each foreground object, 2d opacity is rendered by summing up termination probability $w_{i}$ at each point $x_{i}$ along the ray $r$.
\begin{align}
\hat{O}(r) = \sum_{i=1}^{N}w_{i}
\label{eq: opacity}
\end{align}
Object radiance field is encouraged to be occupied for the pixels within object mask and to be empty for pixels outside object mask. To avoid learning empty signal for occluded parts, \cite{yang2021learning}\cite{kong2023vmap} terminate ray right after it hits the surface of other objects. Both background and foreground objects are supervised by rendered color, depth, and opacity loss.

\textbf{3D learning of object category.}
To learn the geometry and appearance of the category and disentangle shape and texture variations for each object in the category, CodeNeRF consists of two parts. The first part is responsible for geometry, taking 3d point $x$ and shape code $z_{s}$ as input and producing volume density $\sigma$ and feature vector $v$. The second part is responsible for the appearance, taking $v$ and texture code $z_{t}$ as input and outputting RGB color $c$.
\begin{equation}
\begin{split}
    F_{\theta} &: F_{\theta_{s}}^{s}\circ F_{\theta_{t}}^{t} \\
    F_{\theta_{s}}^{s} &: (\gamma_{x}(x),z_{s})\rightarrow (\sigma,v) \\
    F_{\theta_{t}}^{t} &: (v,\gamma_{d}(d),z_{t})\rightarrow c
\end{split}
\end{equation}
\subsection{Category-level registration}
\label{registration}
Learning the shared 3D information of objects within the same category requires aligning these objects in a 3D space accurately. Existing category-level 3D learning methods either operate exclusively on synthetic datasets \cite{jang2021codenerf}\cite{park2019deepsdf}, where 3D points are represented in an object-centric coordinate, or rely on pose from pretrained 3D detection networks \cite{muller2022autorf}\cite{kundu2022panoptic}. Therefore it is difficult to apply these methods for categories with limited 3D data. In typical indoor settings, another challenge arises where objects may be partially observed due to occlusions or a limited number of viewpoints. This makes the alignment of objects into a common coordinate frame difficult. Moreover, objects predicted to be in the same category might have large differences in shape, making training a shared model inefficient. In this section, we address these challenges with the following strategies.

\textbf{Uncertainty guided representative selection.}
We select a well-observed object as a representative for its category and use it in subsequent registration stages. Identifying well-observed objects in cluttered environments, especially with occlusions, is non-trivial. We tackle this by evaluating ray uncertainty in various viewing directions using the NeRF model trained for individual objects. First, we train an object-level model for each object by utilizing our batch version implementation of vMAP\cite{kong2023vmap}. The trained network serves as a compact memory for observations of each object. Inspired by \cite{lee2022uncertainty}, we calculate ray uncertainty by analyzing the weight distribution predicted by the network along the ray. 

Rays that travel the regions accurately learned by the object-level model have a clear peak in weight distribution, while rays that travel poorly learned parts have noisier peaks. The concentration of a weight distribution of ray $r$ can be quantified using entropy $H(r) = -\sum_{i=1}^{N}w_{i}\log(w_{i})$. Additionally, from Eq. \ref{eq: opacity}, the sum of weights can determine whether the ray crosses the empty space only. From these properties, we define a reliability metric $g(u(r))$ as explained below. First, we design function $u$ from weight distribution as:
\begin{align}
    u(r) = \left ( \sum_{i=1}^{N}w_{i} \right ) \cdot \exp(-\alpha \cdot H(r))\;,
\end{align}
where $0 \leq w_{i} \leq 1$, $H(r) \geq 0$, and $0 \leq u(r) \leq 1$.
Rays looking at empty regions have a low $u(r)$, accurate regions have a high $u(r)$, and uncertain regions have a medium $u(r)$. We distinguish uncertain regions using another reliability function $g$ designed as the sum of two sigmoid functions:
\begin{align}
    g(u) = \sigma(-\alpha_{m}(u-\beta_{m})) + \sigma(\alpha_{M}(u-\beta_{M}))
\end{align}
\begin{align*}
    \alpha_{M} &= \frac{2\log(\frac{\kappa}{1-\kappa})}{M_{2}-M_{1}}, \; \alpha_{m}=\frac{2\log(\frac{\kappa}{1-\kappa})}{m_{2}-m_{1}}, \\
    \beta_{M} &= \frac{M_{1}+M_{2}}{2}, \; \beta_{m}=\frac{m_{1}+m_{2}}{2}\;.
\end{align*}
Rays looking at empty regions or accurate regions have a high $g(u(r))$, and uncertain regions have a low $g(u(r))$.

Given the RGB-D sequence and object masks, we acquire the observed point cloud for each object. We then cast rays uniformly from points on a spherical surface, which has a radius 1.2 times the largest dimension of the observed point cloud, to their antipodal points. In this way, we can predict uncertainty for all parts of each object as represented in Fig. \ref{fig:case2}. For each category, the object with the highest percentage of rays of which reliability is above a threshold $\eta$ is chosen as representative. 

\textbf{Registration of objects to the representative.}
To represent various objects within the same category in a consistent object-centric coordinate, we align each object to the representative of its category using point cloud registration algorithms. We use Teaser++ \cite{yang2020teaser} because it robustly aligns objects that look slightly different or are partially observed. After alignment, we adjust oriented bounding box (OBB) according to the aligned pose, resulting the bound being close to ground truth for partially observed objects. The refined OBB is used to map world coordinate to aligned coordinate in unit cube which is called Normalized Object Centric Space (NOCS) \cite{muller2022autorf} in the further training stage.

\textbf{Subcategorization.}
It is unclear whether sharing the shape information is beneficial for the 3D learning of objects that significantly vary in shape despite belonging to the same category. Especially in scenarios like ours, where learning is confined to a limited set of observed objects without prior knowledge, articulating the advantages of training a single model becomes more challenging.

\begin{figure}[t]
\centering
\includegraphics[width = 1\linewidth]{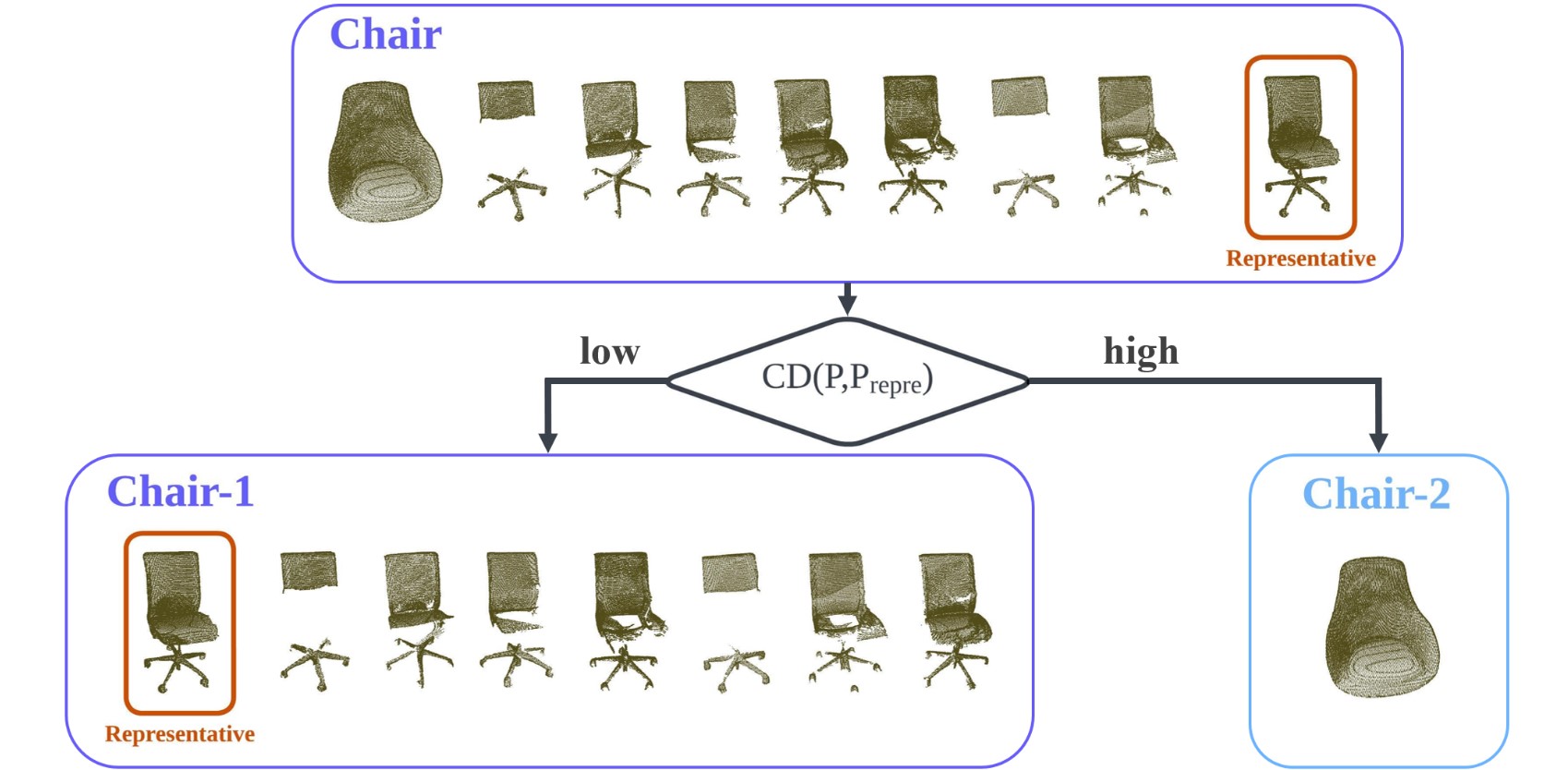}
\caption{A schematic diagram of subcategorization module}
\vspace{-5mm}
\label{fig:figure3_}
\end{figure}

Such a strategy might consume network capacity inefficiently, deteriorating performance compared to individual training. To mitigate this, we conduct subcategorization, as shown in Fig. \ref{fig:figure3_}, based on a simple assumption that objects poorly aligned to each other do not share meaningful information about shape. Thanks to the robustness of our registration module against noise and partial observations, this assumption stands. For quantitative evaluation of registration, we use the unidirectional Chamfer distance $CD_{unidir}(P,Q)=\frac{1}{\left| P\right|}\sum_{p\in P}\text{min}_{q\in Q}\left| p-q\right|$, where $P$ and $Q$ are the point cloud of the object to be aligned and the point cloud of the representative object, respectively. If the Chamfer distance is larger than the threshold, the object is distinguished by a different subcategory from the representative object. We chose \textit{unidirectional} metric because $P$ is often partially observed.

\subsection{Model training}
\label{training}
Given $N$ categories in the scene detected from the sequence, we employ one large model for the background and $N$ smaller models dedicated to each category. Each category-level model follows a structure similar to CodeNeRF. However, as the number of objects trained in each category is generally much fewer than that in CodeNeRF, we utilize a much smaller model. Since our objective leans more towards 3D reconstruction rather than view synthesis, we do not incorporate the viewing direction. Both the background and individual category-level models are trained from scratch. The training of our background model is the same as that of vMAP except that our method is a batch approach. 

In every training iteration of the category-level model, we randomly sample training pixels gathered from 2D bounding boxes of objects in the category. This straightforward approach is enough to let the category-level model learn shape information of all the observed parts of the objects in its category. We sample 10 points along each ray $r$ using depth-guided sampling proposed in vMAP. Each sampled point $^{w}\textrm{x}_{i}$ is mapped to $^{o}\textrm{x}_{i}$ represented in NOCS using pose and 3d bound of its object. The point is then fed into the category-level model, conditioned by the current estimate of the shape and texture code corresponding to the object that the point belongs to. For each input point $^{o}\textrm{x}_{i}$, the model outputs the occupancy probability $o_{i}$ and color $c_{i}$ for each point $^{o}\textrm{x}_{i}$. The termination probability, i.e, the weight $w_{i}$, becomes $w_{i}=o_{i}\prod_{j=1}^{i-1}(1-o_{j})$. Using volume rendering, color, depth, and opacity are rendered in the form of a weighted sum. Training proceeds by object-level supervision of depth, color, opacity, and regularization \cite{jang2021codenerf} derived from the prior of the latent vector.
\begin{align}
    L = \sum_{k=1}^{K}&L_{depth}^{k}+\lambda_{1}L_{color}^{k}+\lambda_{2}L_{opacity}^{k}+\lambda_{3}L_{reg}^{k}\;, \\
    L_{depth}^{k} &= \sum_{r\in R^{k}}M^{k}(r)\left| \hat{D}(r)-D(r)\right|\;,  \\
    L_{color}^{k} &= \sum_{r\in R^{k}}M^{k}(r)\left| \hat{C}(r)-C(r)\right|\;, \\
    L_{opacity}^{k} &= \sum_{r\in R^{k}}\left| \hat{O}(r)-M^{k}(r)\right|\;, \\
    L_{reg}^{k} &= \left\| z_{s}^{k}\right\|_{2}^{2}+\left\| z_{t}^{k}\right\|_{2}^{2}\;,
\end{align}
where $k\left (=1,2,\ldots,K \right)$, $R^{k}$, $M^{k}$ are index, 2d bounding box, and 2d object mask of the object, respectively.
\begin{figure}[t]
\centering
\includegraphics[width = 1\linewidth]{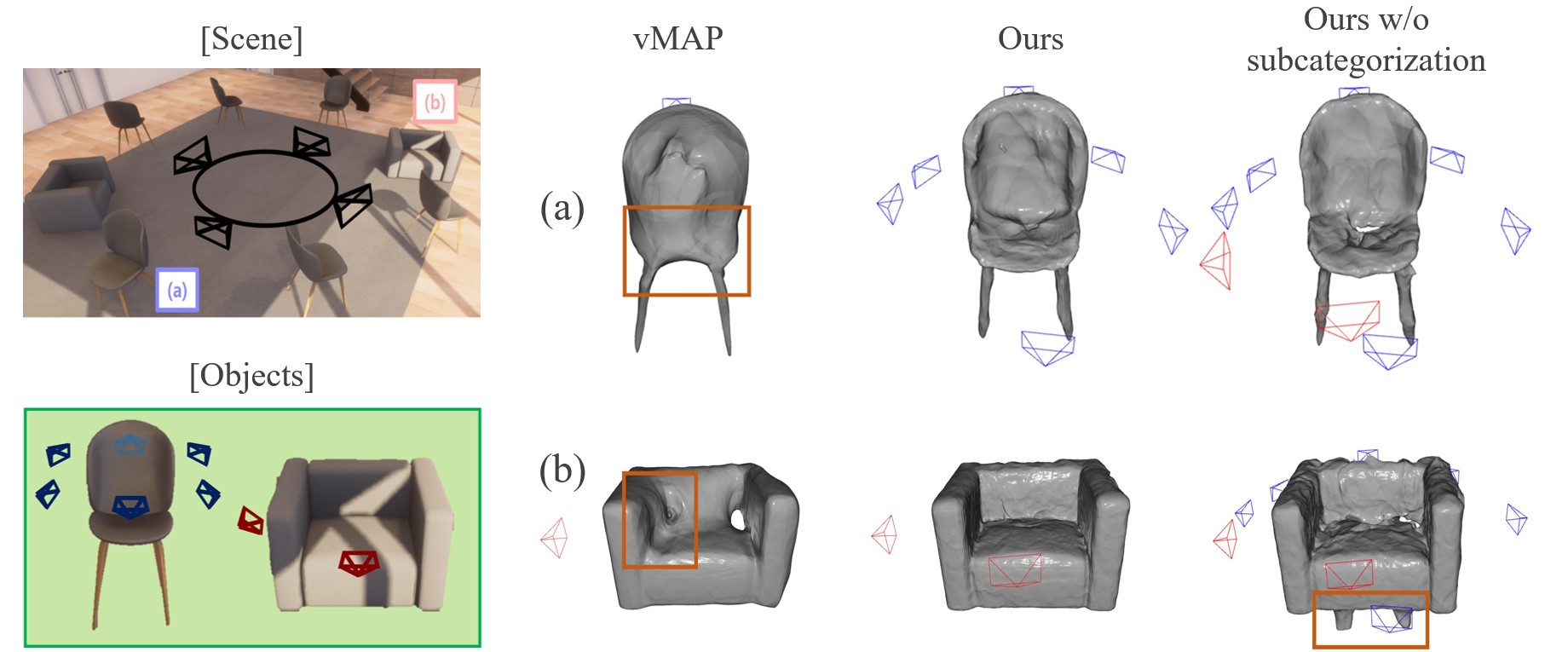}
\caption{Reconstruction of unobserved region}
\vspace{-5mm}
\label{fig:case1}
\end{figure}

\section{EXPERIMENT}
We evaluate our object-level mapping system both qualitatively and quantitatively on synthetic and real-world datasets, comparing it with prior state-of-the-art neural implicit scene and object reconstruction methods. We thoroughly assess the role of each component in our system through ablation study.

\subsection{Experiment setup}
\textbf{Implementation details.}
All experiments are conducted using an NVIDIA RTX A5000 GPU. We compare our method with the most closely related method, vMAP. For a fair comparison with vMAP, we modify vMAP from its original online format to a batch version, denoted as vMAP*, and set hyperparameters that determine the trade-off between quality and efficiency as follows. In every iteration, we sample $120\times \frac{n_{obj}}{n_{cls}}$ and $120$ pixels for each category-level model of our method and instance-level model of vMAP*, respectively. Both methods utilize 4-layer MLPs with hidden dimensions of 128 for the background and 32 for the foreground. Our category-level model adopts architecture similar to CodeNeRF to incorporate shape and texture codes, each with a dimension of 32. In this way, we utilize more network parameters while maintaining a similar or smaller number of total parameters per test scene. We set $1\times 10^{4}$ and $1.5\times 10^{4}$ for our method and vMAP*, respectively. Both methods take 20-30 minutes for training.

For training speed, we adopt vMAP's vectorized training scheme for category models. We utilize object masks that maintain temporal consistency between frames. These are either provided by the dataset or, in the case where object masks are not provided, can be obtained by leveraging the semantic and spatial consistency mentioned in vMAP for association. 

For the registration process, we employ the official Teaser++ Python code, executing under the simultaneous pose and correspondence (SPC) setting in the paper. To enhance registration robustness, the template undergoes transformation to one of 24 possible initial poses defined by the OBB that fits the template. The alignment that results in the smallest unidirectional Chamfer distance is selected. 
In the subcategorization process, the unidirectional Chamfer distance is normalized using the largest dimension of the template point cloud to ensure consistent thresholds across various category scales. 

\begin{figure}[t]
\centering
\includegraphics[width = 1\linewidth]{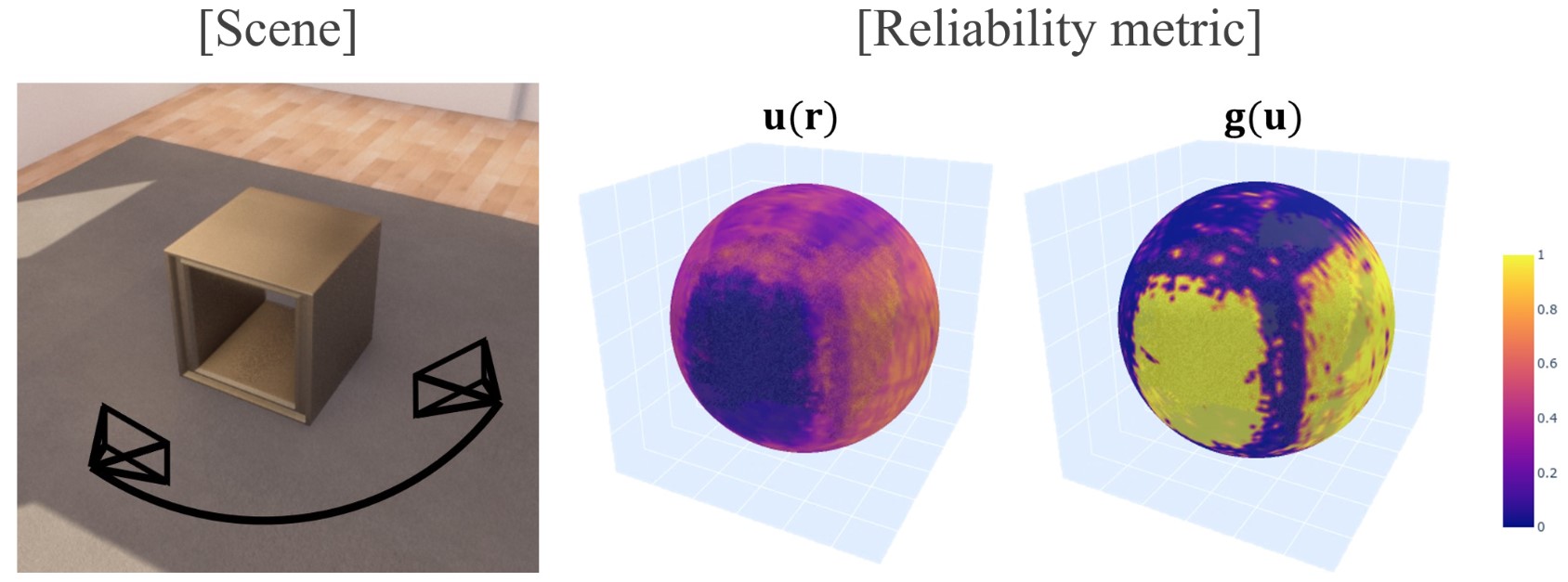}
\caption{Visualization of proposed reliability metric. Color value in each spherical surface point indicates $u(r)$ and $g(u)$ value of the point's corresponding ray. Both plots are oriented as same as the object in the scene.} 
\vspace{-5mm}
\label{fig:case2}
\end{figure}

All the experiments use the same hyperparameters as follows: threshold in representative selection $\eta$ is set to be 0.5. In the reliability metric, as shown in Fig. \ref{fig:case2}, three types of regions (i.e., well observed, unobserved, empty) have different range of $u(r)$, so we set $m_{1}=0.1$, $m_{2}=0.15$, $M_{1}=0.57$, $M_{2}=0.65$, and $\kappa_{1}=0.9$ following the transition region for $u(r)$. Especially, we set the threshold for subcategorization $\eta$ as 0.12. Note that we select this value conservatively low to avoid the severe case where very different objects are misclassified into the same subcategory and damage their reconstruction. The loss weights are set at $\lambda_{1}=5$, $\lambda_{2}=10$, and $\lambda_{3}=0.0005$.

\begin{table*}[t]
\centering
\caption{Object-level reconstruction results on 8 Replica scenes}
\label{tab: table1}
\resizebox{.82\textwidth}{!}{%
\begin{tabular}{llccccccccc}
\hline
                                       &                                              & room-0         & room-1         & room-2         & office-0       & office-1       & office-2       & office-3       & office-4           \\ \hline
\multirow{3}{*}{\textbf{TSDF-Fusion*}} & \textbf{Acc.}                       & 2.80           & 3.00           & 4.10           & 3.19           & 2.44           & 2.94           & 3.21           & 3.75           \\
                                       & \textbf{Comp.}                      & 3.79           & 5.06           & 4.23           & 3.08           & 4.45           & 4.17           & 3.63           & 3.65           \\
                                       & \textbf{C.R.} & 77.22          & 73.36          & 77.75          & 83.45          & 80.48          & 81.45          & 80.26          & 80.68          \\ \hline
\multirow{3}{*}{\textbf{iMAP*}}        & \textbf{Acc.}                       & 3.16           & 3.30           & 4.36           & 3.48           & 2.64           & 3.12           & 3.68           & 4.06           \\
                                       & \textbf{Comp.}                      & 2.79           & 4.16           & 3.89           & 3.31           & 3.32           & 3.13           & 3.64           & 4.36           \\
                                       & \textbf{C.R.} & 84.17          & 76.62          & 79.37          & 82.08          & 84.24          & 82.15          & 80.65          & 77.11          \\ \hline
\multirow{3}{*}{\textbf{ObjectSDF++*}} & \textbf{Acc.}  & 2.27   & 3.99   & 2.34   & 2.35     & 2.70     & 2.32     & 2.28     & 2.19     \\
                                        & \textbf{Comp.} & 3.29   & 5.51   & 4.42   & 3.13     & 3.75     & 3.72     & 3.65     & 6.36     \\
                                        & \textbf{C.R.}    & 83.53  & 71.13  & 79.21  & 84.68    & 81.18    & 83.57    & 83.13    & 74.60  \\ \hline
\multirow{3}{*}{\textbf{vMAP*}}        & \textbf{Acc.}                       & \textbf{1.98}           & \textbf{2.73}           & 2.19           & \textbf{2.17}           & \textbf{2.19}           & 2.16           & 2.28           & \textbf{2.08}           \\
                                       & \textbf{Comp.}                      & 1.70           & 2.55           & 2.61           & 1.23           & 3.18           & 2.05           & 2.08           & 2.11           \\
                                       & \textbf{C.R.} & 94.80          & 90.54          & 86.83          & 95.95          & 90.08          & 93.97          & 92.28          & 90.57          \\ \hline
\multirow{3}{*}{\textbf{Ours}}         & \textbf{Acc.}                       & 2.21           & 3.16           & \textbf{2.15}  & 2.26           & 2.22           & \textbf{2.13}  & \textbf{2.23}  & \textbf{2.08}           \\
                                       & \textbf{Comp.}                      & \textbf{1.67}  & \textbf{2.24}  & \textbf{1.92}  & \textbf{1.22}  & \textbf{3.15}  & \textbf{1.89}  & \textbf{1.33}  & \textbf{1.83}  \\
                                       & \textbf{C.R.} & \textbf{94.83} & \textbf{92.26} & \textbf{89.67} & \textbf{96.01} & \textbf{90.31} & \textbf{94.64} & \textbf{95.89} & \textbf{92.57} \\ \hline
\end{tabular}%
}
\end{table*}
\begin{figure*}[t]
\centering
\includegraphics[width = 1\linewidth]{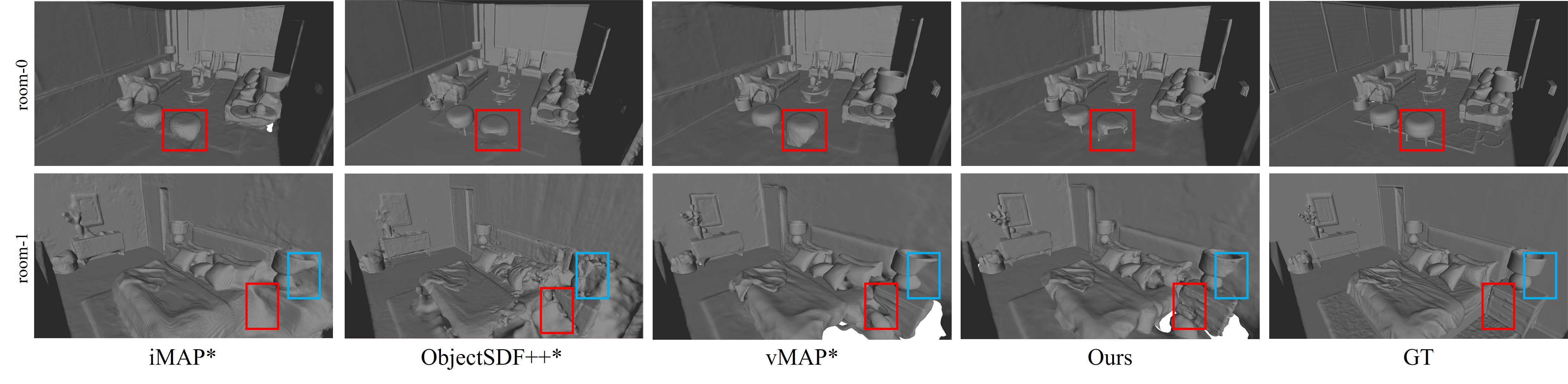}
\caption{Reconstruction results of Replica scenes.
Especially when similar objects are present, our method better reconstructs unseen parts for various categories (ottoman, table, nightstand, lamp) compared to baselines.}
\vspace{-5mm}
\label{fig:replica}
\end{figure*}
\textbf{Datasets.}
We evaluate the proposed system on Replica \cite{straub2019replica} and ScanNet \cite{dai2017scannet} datasets. Replica is a synthetic dataset that provides RGB images along with ground truth (gt) poses, gt depths, and gt object/semantic masks. ScanNet is a real-world dataset that provides RGB images accompanied by gt poses, noisy depths, and noisy object/semantic masks. In experiments on Replica, we use 2000 frames of 8 room-scale scenes as utilized in iMAP \cite{sucar2021imap}. In ScanNet dataset, we select 6 scenes. It should be noted that object point clouds and bounds derived from ScanNet's noisy object masks are inaccurate due to depth discontinuities at object boundaries. Thus, as preprocessing, we generate semantically refined depth segmentation masks, as proposed in \cite{grinvald2019volumetric}, and use them instead of the noisy object masks.

\textbf{Metrics.}
For both datasets and ablation studies, we employ object-level metrics as same as vMAP except completion ratio (\textless $1$ cm $\%$): accuracy [cm] (noted as Acc.), completion [cm] (noted as Comp.), and completion ratio (\textless $5$ cm $\%$) (noted as C.R.). Since the accuracy measures the average error of the reconstructed parts, inferring details of unseen parts using category-level shape information might incur a metric penalty. Therefore, we evaluate accuracy for the same part of the resulting mesh of each method by cutting off using 3d OBB fitted from vMAP mesh, for fairer comparison. 

\subsection{Case Study}
To understand how the category-level model and subcategorization in our method improve the reconstruction of unobserved parts for objects, we test the reconstruction of unobserved parts using the environment in Fig. \ref{fig:case1}. To focus on the impacts of the proposed modules, we use the ground truth pose for each object. The test environment consists of two significantly different types of chair instances where different parts of each instance are observed. For each method, two selected instances from each chair type are visualized with camera frames used for training the model that is responsible for predicting the shape of the instance. Each frame relates to a different instance, observing the instance the best in the sequence. vMAP cannot reconstruct the seat of instance (Fig. \ref{fig:case1}(a)) and oversmooths the unobserved part (Fig. \ref{fig:case1}(b)). The category-level model without subcategorization reconstructs the unobserved part in Fig. \ref{fig:case1}(a),  but wrongly reconstructs in Fig. \ref{fig:case1}(b) by generating the leg from the shape information of instances in a different type. Only our method can accurately reconstruct the unobserved part, since it takes additional information from only chair instances with the same type. 

In Fig. \ref{fig:case2}, we also visualize the reliability metric to check whether it plays an appropriate role. The result meets well our expectations: the observed regions as vacancy and well-observed areas have high $g(u(r))$ values, whereas the unobserved regions show low $g(u(r))$ values.

\subsection{Evaluation on Scene and Object Reconstruction}
\textbf{Results on Replica dataset.}
Table \ref{tab: table2} shows the object-level reconstruction results. Our method is compared with scene-level methods (i.e., TSDF-Fusion and iMAP) and object-level methods (i.e., ObjectSDF++ and vMAP). Since these methods except ObjectSDF++ \cite{wu2023objectsdf++} are online methods, we implement their batch mode using their open-source code and train them using the ground-truth pose (denoted with ``*") for a fair comparison. Since ObjectSDF++ uses the predicted depth from [38], we re-implement it to use the ground-truth depth and replace its scale-invariant depth loss with the usual scale-aware depth loss. We train the re-implemented ObjectSDF++ with the same iteration number as the original ($2\times 10^5$ iteration, 18 hours).

Our method outperforms the baselines in object-level completion and completion ratio for all the scenes and shows comparable or better accuracy than baselines. Ours and vMAP achieve significantly higher completeness than ObjectSDF++ because they show better hole-filling capability of unobserved regions than ObjectSDF++. It is because they explicitly avoid providing training information about occluded regions to their models, whereas ObjectSDF++ does not. Our method achieves better completeness than vMAP* since ours reconstructs unobserved parts more accurately using category-level information. Notably, in scenes like room-2 and office-3, where there are many occlusions and several objects of the same shape, our method improves object-level completion by 20-30$\%$ compared to vMAP. Fig. \ref{fig:replica} shows the reconstruction results of 2 Replica scenes. Highlighted boxes represent parts that are not visible in the given sequence, and our method outperforms the baselines in reconstructing details of these parts.

\begin{table}[]
\centering
\caption{Object-level reconstruction results on ScanNet.
}
\label{tab: table2}
\resizebox{\columnwidth}{!}{%
\begin{tabular}{llcccccc}
\hline
                                & scene         & 0013           & 0024           & 0059           & 0066           & 0084           & 0281           \\ \hline
\multirow{3}{*}{\textbf{vMAP*}} & \textbf{Acc.}  & 4.50           & 3.74           & 4.53           & 4.44           & 2.91           & 4.98           \\
                                & \textbf{Comp.} & 3.62           & \textbf{3.98}  & 4.57           & 4.50           & 4.58           & 4.09           \\
                                & \textbf{C.R.}  & 89.05          & \textbf{84.63} & 80.05          & 81.33          & 86.11          & 84.20          \\ \hline
\multirow{3}{*}{\textbf{Ours}}  & \textbf{Acc.}  & \textbf{3.61}  & \textbf{3.43}  & \textbf{4.23}  & \textbf{4.10}  & \textbf{2.82}  & \textbf{4.56}  \\
                                & \textbf{Comp.} & \textbf{3.58}  & 4.19           & \textbf{4.31}  & \textbf{4.13}  & \textbf{4.06}  & \textbf{3.84}  \\
                                & \textbf{C.R.}  & \textbf{89.53} & 84.07          & \textbf{81.29} & \textbf{82.09} & \textbf{86.79} & \textbf{85.43} \\ \hline
\end{tabular}%
}
\vspace{-2mm}
\end{table}
\begin{figure}[t]
\centering
\includegraphics[width = .9\linewidth]{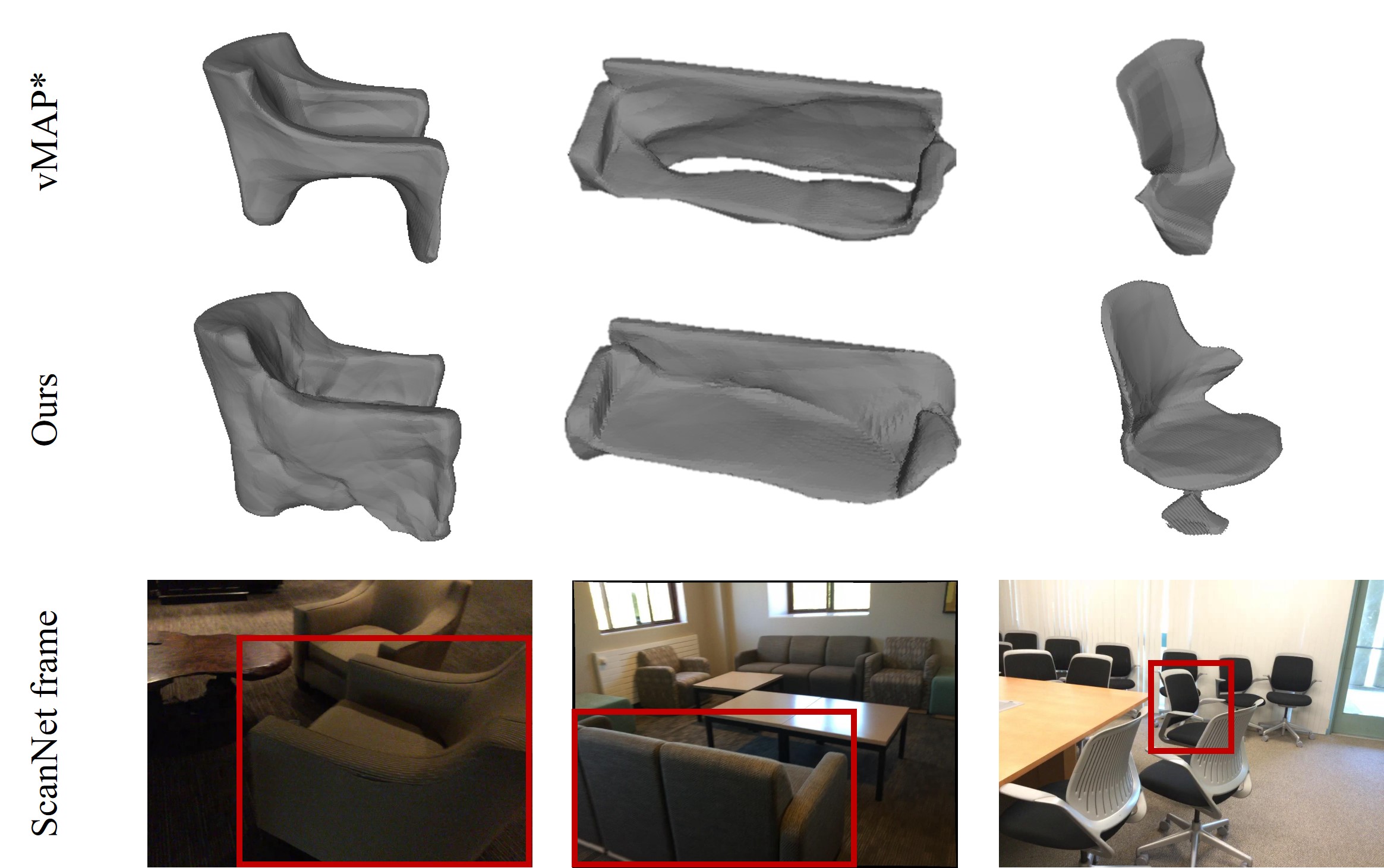}
\caption{Object-level reconstruction results on ScanNet.} The bottom row shows examples of frames that capture the corresponding object.
\vspace{-5mm}
\label{fig:scannet}
\end{figure}
\textbf{Results on ScanNet dataset.}
To highlight the strengths and practicality of our system, we conduct experiments on 4 scenes - scene 0013/0059/0066/0281 - that contain several similar objects and 2 scenes that contain a small number of similarly shaped objects and  from the real-world ScanNet dataset. Table \ref{tab: table3} compares the reconstruction results of our approach with vMAP for all foreground objects in the ScanNet scenes. Our method outperforms vMAP in all metrics for 4 scenes with multiple similar objects, and shows comparable performance to vMAP* in 2 scenes with a few similar objects. As shown in Fig. \ref{fig:scannet}, our method reconstructs the objects more completely.

\textbf{Category-wise results.} We compare our method with vMAP* for objects in certain categories to clarify the strength of our method in Table \ref{tab: table3}. In both datasets, our method demonstrates significant enhancements over vMAP* for multiple objects that belong to the same class. Even for objects without similar ones in the scene, which is not our target, our method at least performs similarly to vMAP.

\begin{table}[t]
\centering
\caption{Object-level reconstruction results for selected classes. room-2 contains multiple chairs and a single box, and scene0066 contains multiple chairs and a single table.}
\label{tab: table3}
\resizebox{0.8\columnwidth}{!}{%
\begin{tabular}{llcccc}
\hline
                                &                         & \multicolumn{2}{c}{room-2} & \multicolumn{2}{c}{scene0066} \\ \hline
                                &                         & chair            & box             & chair             & table             \\ \hline
 & \textbf{Acc.}  & 3.67             & 0.87            & 4.96              & \textbf{5.41}     \\
                              {\textbf{vMAP*}}  & \textbf{Comp.} & 4.83             & 0.89            & 6.89              & \textbf{1.99}     \\
                                & \textbf{C.R.}    & 78.82            & 100             & 70.65             & \textbf{96.90}    \\ \hline
 & \textbf{Acc.}  & \textbf{3.42}    & \textbf{0.81}   & \textbf{4.58}     & 5.44              \\
                              {\textbf{Ours}}  & \textbf{Comp.} & \textbf{2.03}    & \textbf{0.81}   & \textbf{5.77}     & 2.02              \\
                                & \textbf{C.R.}    & \textbf{94.48}   & \textbf{100}    & \textbf{73.56}    & 96.79             \\ \hline
\end{tabular}%
}
\end{table}
\begin{table}[]
\centering
\caption{Object-level reconstruction results of ablation study. The performances of representative selection and subcategorization methods are computed using the categories with more than 5 and 2 objects included, respectively.}

\label{tab: table4}
\resizebox{.85\columnwidth}{!}{%
\begin{tabular}{|ll|c|c|c|}
\hline
\multicolumn{2}{|c|}{}                                            & \textbf{Acc.}          & \textbf{Comp.}          & \textbf{C.R.}            \\ \hline
\multicolumn{1}{|l|}{\multirow{2}{*}{Representative}}    & Random & 1.83          & 2.17          & 91.59          \\
\multicolumn{1}{|l|}{}                                   & Ours   & \textbf{1.82} & \textbf{1.87} & \textbf{92.70} \\ \hline
\multicolumn{1}{|l|}{\multirow{2}{*}{Subcategorization}} & No     & 2.42          & 2.24          & 91.83          \\
\multicolumn{1}{|l|}{}                                   & Ours   & \textbf{2.14} & \textbf{2.07} & \textbf{92.77} \\ \hline
\end{tabular}
}
\end{table}
%
%
%
%
\begin{figure}[b]
\vspace{-3mm}
\centering
\includegraphics[width = .9\linewidth]{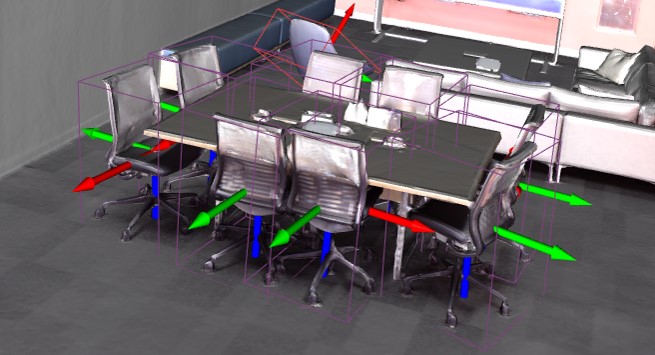}
\caption{Category-level registration result. OBB, 6DoF pose, and subcategory for each chair instance are visualized. Note that the chair class is colored by two different colors, which means that the chair class is divided into two subcategories.}
\label{fig:pose&bound}
\end{figure}
\subsection{Ablation study}
We compare the 3D reconstruction performance between the proposed uncertainty-guided method and simple random selection for selecting representative objects. Table \ref{tab: table4} indicates that, especially in scenarios with many objects belonging to the same category where choosing an appropriate representative is vital, both accuracy and completion are superior when the proposed method is utilized. Table \ref{tab: table4} also denotes that both accuracy and completion are superior when subcategorization is applied compared to when it is not. 

In addition, Fig. \ref{fig:pose&bound} displays the estimated poses and bounds of the observed objects that belong to a selected category (chair) along with subcategorization outcomes, as the result of category-level registration. Our method can estimate the bounds and poses of chairs that are occluded by table consistently with other objects in the same category. We also compare the memory usage with vMAP*. Our single category-level model has 18179 parameters, whereas vMAP*'s single instance-level model has 11363 parameters. Note that our model can learn all objects in the category using a smaller amount of memory than vMAP*'s several instance-level models, making our system much more memory-efficient in scenes with many objects in the same category.
\section{CONCLUSIONS}
\label{conclusions}
We proposed an object-level mapping system with category-level neural fields. The objective of our system is to enhance the reconstruction of unobserved regions using category-level models without any information learned from external data. To achieve this, we introduce a category-level registration module that maps objects in the same category to the same normalized object-centric space and subcategorizes objects to enforce the category-level model to learn only objects that share meaningful information about shape. Then the model learns neural fields with independent shape and appearance components for each object.
Experiments on synthetic and real-world datasets show that the shape information of objects with similar shapes can be successfully leveraged to reconstruct a more complete mesh.






\bibliographystyle{./bibtex/IEEEtran}
\bibliography{./bibtex/main.bib}

\end{document}